# Statistical Firefly Algorithm for Truss Topology Optimization

Nghi Huu Duong[a], Duy Vo[b,c], Pruettha Nanakorn[a]

[a]*School of Civil Engineering and Technology, Sirindhorn International Institute of Technology, Thammasat University, Pathumthani, Thailand*

[b]*Duy Tan Research Institute for Computational Engineering (DTRICE), Duy Tan University, Ho Chi Minh City, Vietnam*

[c]*Faculty of Civil Engineering, Duy Tan University, Da Nang, Vietnam*

**ABSTRACT**

This study proposes an algorithm titled a statistical firefly algorithm (SFA) for truss topology optimization. In the proposed algorithm, historical results of fireflies' motions are used in hypothesis testing to limit the motions of fireflies that are suggested by current information exchanges between fireflies only to those that are potentially useful. Hypothesis testing is applied to the mechanism of an ordinary firefly algorithm (FA) without changing its structure. As a result, the implementation of the proposed algorithm is simple and straightforward. Limiting the motions of fireflies to those that are potential useful results in reduction of firefly evaluations, and, subsequently, reduction of computational efforts. To test the validity and efficiency of the proposed algorithm, it is used to solve several truss topology optimization problems, including some benchmark problems. It is found that the added statistical strategy in the SFA significantly enhances the performance of the original FA in terms of computational efforts while still maintains the quality of the obtained results.

*E-mail addresses*: nghihd.ce@gmail.com (N. H. Duong); voduy3@duytan.edu.vn (D. Vo); nanakorn@siit.tu.ac.th (P. Nanakorn)

## 1. Introduction

Metaheuristic optimization methods [1, 2] are becoming popular among researchers in the field of structural optimization [3, 4] due to their advantages over traditional optimization methods. Although optimization methods, including metaheuristic methods, have been extensively studied during the recent decades, the studies still leave many things to be desired. This is because real-life problems of structural optimization are large and complex while most of the existing optimal algorithms can handle only small and simple problems. Metaheuristic methods, such as differential evolution (DE) [5-8], genetic algorithms (GAs) [9-11], particle swarm optimization (PSO) [12-15], firefly algorithms (FAs) [16-19], and teaching-learning-based optimization (TLBO) [20-23], provide some degrees of improvement over traditional methods, with respect to abilities to find global optimal solutions. Metaheuristic methods use many search points to perform their searches. As a result, it is more difficult for the searches to be trapped in local optimal regions. In contrast to traditional methods, metaheuristic methods do not require the gradients of objective and constraint functions, which are oftentimes unavailable or computationally expensive to obtain. Among these metaheuristic methods, FAs have been proven to be robust and reliable methods despite their modest applications, when compared to those of other algorithms [24].

Truss optimization problems can be classified into three types, namely size, shape, and topology optimization [25-27]. In size optimization [28-30], the cross-sectional areas of truss members are the design variables, and the coordinates of nodes and the connectivities are held constant. In shape optimization [31-33], the coordinates of nodes become the design variables, while the cross-sectional areas of truss members and the connectivities remain constant. In topology optimization [34-36], the connectivities of truss members are the design variables and they are usually defined using the cross-sectional areas of truss members. As a result, truss topology optimization intrinsically incorporates size optimization. When solving a truss topology optimization problem, the search space that possesses all possible trusses must be defined. The ground structure approach is the most popular scheme for defining search spaces for truss topology optimization problems. In this approach, a search space is defined by using connections between predefined nodes as possible positions of truss elements. The positions of nodes and choices of connections can be carefully selected to manipulate the final search space. One significant disadvantage of the ground structure approach is that the size of a search space increases rapidly when the number of nodes is increased [25, 37, 38]. When the number of nodes is increased, the number of search dimensions increases. Optimization problems with large numbers of search dimensions tend to also have large numbers of local optimal solutions and, as a result, they are difficult to solve. These problems unavoidably require large computational resources.

To avoid being trapped in local optimal regions, robust metaheuristic methods, including FAs, must allow search points to exchange their information as much as possible. Actions suggested by these information exchanges usually require high computational resources. Naturally, it is desirable that only useful actions are performed. In the mechanism of FAs, each firefly must communicate with all other fireflies in the swarm during its moves. These information exchanges do not always direct the fireflies to better locations while time-consuming appraisals of those fireflies that move are always needed. Appraisals of design solutions in structural optimization always need finite element analysis (FEA) which is usually resource-intensive. Therefore, it is beneficial if only those actions that potentially result in better positions of fireflies are executed.

In this study, a statistical firefly algorithm (SFA) is proposed. The proposed algorithm uses mean hypothesis testing [39-41] that utilizes the statistics of past exchanges to predict the usefulness of impending actions that are suggested by current information exchanges. By using hypothesis tests, the effectiveness of actions that are suggested by information exchanges between firefly couples can be estimated. Any actions suggested by firefly couples, whose collaborations are regarded as ineffective by hypothesis tests, will not be executed. This strategy can decrease the computational efforts of traditional FAs particularly in large problems of truss topology optimization using the ground structure approach.

## 2. Hypothesis testing

A hypothesis in statistical analysis [42, 43] is simply a claim or statement about population parameters such as the population mean, variance, and proportion. In inferential statistics, samples are studied instead of the entire population. The obtained results are, nevertheless, used to generalize the entire population. In this study, hypothesis testing of the population mean is used in the proposed FA. In hypothesis testing, two completing hypotheses are considered, namely the null hypothesis and the alternative hypothesis, which is the complement of the null hypothesis. When a hypothesis test is conducted, it is primarily to decide whether to accept or reject the null hypothesis. If the null hypothesis is rejected, it is sufficient to support the alternative hypothesis. On the other hand, if the null hypothesis is not rejected, it is not sufficient to support the alternative hypothesis. Hypothesis tests can be classified into two categories, namely two-tailed and one-tailed hypothesis tests. Let $\mu$ denote the population mean and $\mu_0$ denote a certain value. In two-tailed hypothesis testing, illustrated in Fig 1, the null hypothesis ($H_0$) states that $\mu = \mu_0$. Consequently, the alternative hypothesis ($H_a$) states that $\mu \neq \mu_0$. One-tailed hypothesis testing, shown in Fig 2, consists of two different cases, i.e. left-tailed and right-tailed hypothesis testing. In left-tailed testing, the null hypothesis ($H_0$) states that $\mu \geq \mu_0$ and the alternative one ($H_a$) states that $\mu < \mu_0$. In right-tailed testing, the null hypothesis ($H_0$) states that $\mu \leq \mu_0$ and the alternative one ($H_a$) states that $\mu > \mu_0$.

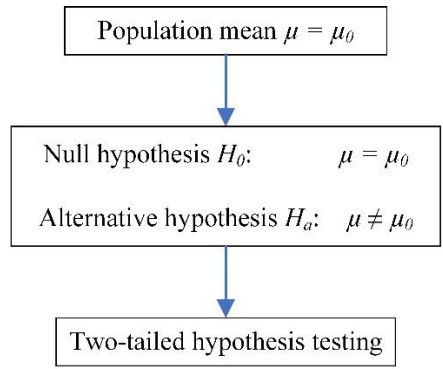

Fig 1. Two-tailed hypothesis testing of the mean

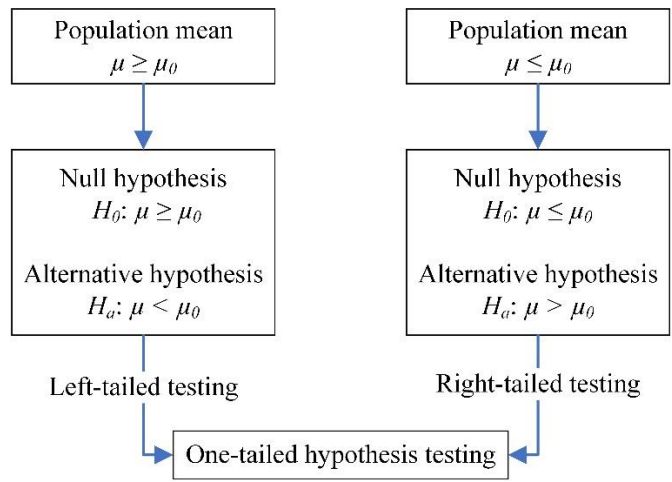

Fig 2. One-tailed hypothesis testing of the mean

The probability of rejecting the null hypothesis when it is true is called the significant level $\alpha$. The significant level $\alpha$ determines the size of the critical or rejection region, in which the null hypothesis is rejected, in the probability distribution curve. The complement of the critical region is called the acceptance region. In a two-tailed hypothesis test, the critical region is two-sided. Each side occupies an area of $\alpha/2$ in the probability distribution curve, as illustrated in Fig 3. In a one-tailed hypothesis test, the critical region, having an area of $\alpha$, can be on the left or right side of the probability distribution curve, as shown in Fig 4. In addition, the values that separate the critical region from the acceptance region are called the critical values. The critical values can be determined when the sample size $n$ and the significant level $\alpha$ are known.

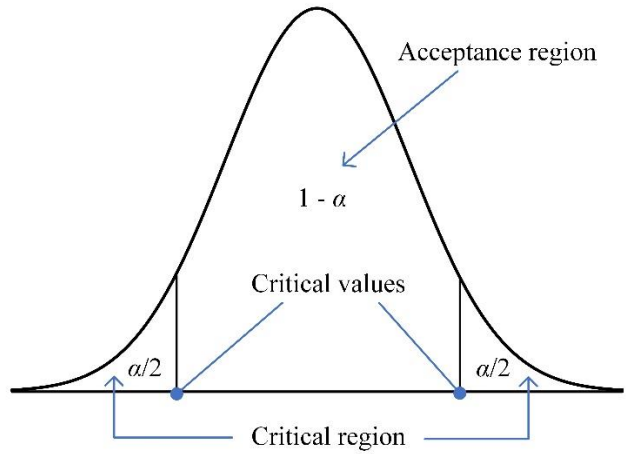

Fig 3. Critical region in two-tailed hypothesis testing

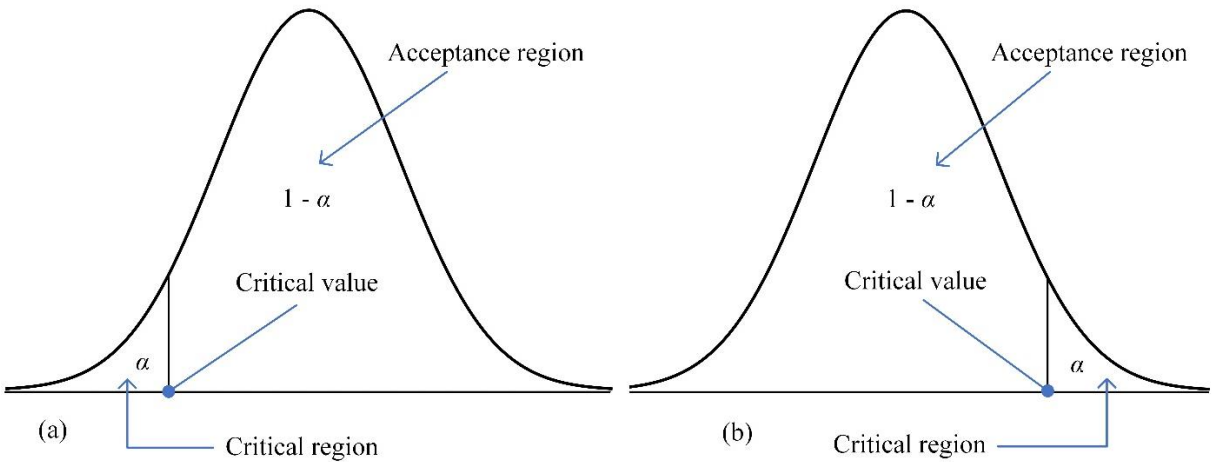

Fig 4. Critical regions in (a) left-tailed hypothesis testing and (b) right-tailed hypothesis testing

*2.2. Normal distribution and t-distribution*

In statistics, t-distributions [44] are used to describe the probability distributions of small samples of normally distributed populations, and they converge into the normal distribution when the sample sizes are large. In this study, the t-distribution is used in hypothesis testing. To determine the critical values, the sample size $n$ and the significance level $\alpha$ are required. Fig 5 shows the difference between the normal distribution and a t-distribution. When a t-distribution is used, the critical values are $-t_{\alpha/2}^{(n-1)}$ and $t_{\alpha/2}^{(n-1)}$ in two-tailed hypothesis testing. In addition, the critical values are $-t_{\alpha}^{(n-1)}$ and $t_{\alpha}^{(n-1)}$ for left-tailed and right-tailed hypothesis testing, respectively.

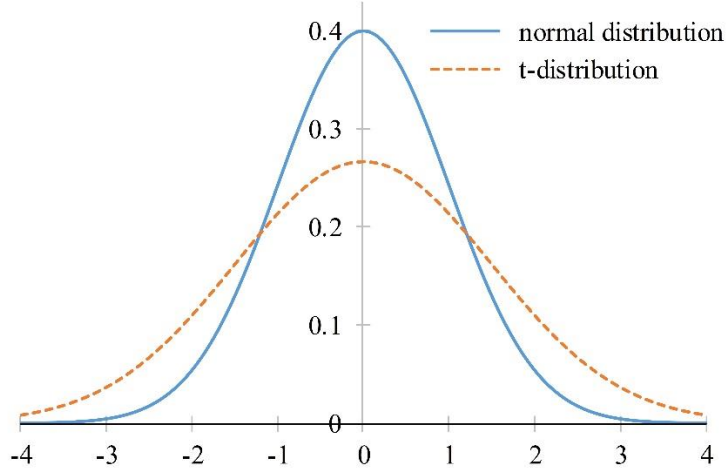

Fig 5. Normal distribution and t-distribution

*2.3. Statistical formulation*

The test statistic computed from a t-distribution for hypothesis testing is denoted by $t$ and expressed as

$$t = \frac{\bar{m} - \mu_0}{\frac{s}{\sqrt{n}}} \quad (1)$$

where $\bar{m}$ is the sample mean and $s$ is the sample standard deviation. The null hypothesis is rejected when $t$ falls in the critical region. For two-tailed hypothesis testing, the null hypothesis is rejected when the value of $t$ is in $\left(-\infty, -t_{\alpha/2}^{(n-1)}\right)$ or $\left(+t_{\alpha/2}^{(n-1)}, +\infty\right)$. For left-tailed and right-tailed hypothesis testing, the null hypothesis is rejected when the value of $t$ is in $(-\infty, -t_{\alpha}^{(n-1)})$ and $(+t_{\alpha}^{(n-1)}, +\infty)$, respectively.

**3. Firefly algorithms**

FAs are nature-inspired algorithms [16, 45] that mimic the social behavior of fireflies. The algorithms are based on the following concepts:

Between any two fireflies, the worse one will move towards the better one. A firefly will move randomly if there is no better one. The merit of a firefly is determined based on the objective and constraints of the problem.

How far the worse firefly will move towards the better one depends on the attractiveness of the better one. The attractiveness of a firefly decreases as the distance to the firefly increases.

The attractiveness $\beta$ determines how strong a firefly attracts other fireflies, and is defined as [46]

$$\beta(r) = \beta_0 e^{-\gamma r^2} \tag{2}$$

where $r$ is the distance to the firefly and $\beta_0$ is the attractiveness at $r = 0$. Moreover, $\gamma$ is a fixed light absorption coefficient.

Conventionally, the position of firefly $i$, attracted by another brighter firefly $j$, is updated by the following update equation, i.e.

$$x_{i,k} = x_{i,k} + \beta(r_{ij})(x_{j,k} - x_{i,k}) + \varepsilon\left(rnd_{i,k} - \frac{1}{2}\right). \tag{3}$$

Here, $x_{i,k}$ and $x_{j,k}$ are the $k^{th}$ components of fireflies $i$ and $j$, respectively. In addition, $r_{ij}$ is the distance between the two fireflies. The third term on the right-hand side is a randomization term, where $\varepsilon$ is a coefficient and $rnd_{i,k}$ is a random number on [0,1].

Traditional FAs have been further modified in many research works [47-51] in order to adapt to a variety of real-life problems. In this study, the update equation is a hybrid version modified from the versions in [45] and [47], given by

$$x_{i,k} = x_{i,k} + \beta(\hat{r}_{ij})rnd_{i,k}(x_{j,k} - x_{i,k}) + \varphi_{i,k}\omega^\tau\left(rnd_{i,k} - \frac{1}{2}\right). \tag{4}$$

The above equation is originally based on the work [45]. The second and the third components on the right-hand side are, respectively, the terms of attractiveness and randomization. In this study, the term of attractiveness is multiplied by a random number on [0,1]. In addition, the distance between the two fireflies is replaced by a normalized distance $\hat{r}_{ij}$ [47], given by

$$\hat{r}_{ij} = \frac{r_{ij}}{r_{max}}. \tag{5}$$

Here, $r_{max}$ is the maximum possible distance between any two fireflies, given by

$$r_{max} = \sqrt{\sum_{k=1}^{D}(up_k - low_k)^2} \tag{6}$$

where $D$ is the number of design variables. In addition, $up_k$ and $low_k$ are, respectively, the upper and lower bounds of the $k^{th}$ design variable. The normalized distance $\hat{r}_{ij}$ is necessary because in a high-dimensional and large-sized problem, the fireflies in the swarm can be far away from each other. Since the attractiveness is inversely proportional to the distance between each pair of fireflies, the algorithm can perform poorly when the unnormalized distances are used. Moreover, $\omega$ in Eq. (4) is a cooling factor [45] and $\tau$ denotes the current iteration number. Lastly, $\varphi_{i,k}$ is the initial randomness scaling factor [45], which is expressed as

$$\varphi_{i,k} = \theta_{i,k}(up_k - low_k). \tag{7}$$

Here, $\theta_{i,k}$ is a user-defined coefficient, which, in this study, is a random number on [0,1]. Note that developing a new FA is not the objective of this study. In fact, any good version of FAs can be employed.

## 4. Statistical firefly algorithm

In an FA, firefly $i$ compares itself with firefly $j$. If firefly $j$ is better than firefly $i$, firefly $i$ moves towards firefly $j$. This process represents collaborations between the two fireflies. When firefly $i$ moves towards a better firefly $j$, there is no guarantee that firefly $i$ will become better. In this study, the quality of the collaborations between firefly couples are collected continuously by considering the results of their movements from the collaborations. For every move of firefly $i$ towards firefly $j$, if the move results in a better location of firefly $i$, a score of 1 is collected. In contrast, if a move does not result in a better location of firefly $i$, a score of 0 is collected. To be able to start hypothesis testing, it is assumed that, before the beginning of the algorithm, each firefly $i$ has already had a successful imaginary move to firefly $j$, and, as a result, a score of 1 is initially collected and the initial number of moves is equal to 1. After only one real move, the mean and standard deviation of the collaboration scores can then be computed. If the mean is high, it means that the collaborations of firefly $i$ with firefly $j$ are usually useful, and it is likely that the future collaborations of firefly $i$ with firefly $j$ will be useful as well.

In this study, to accept that a collaboration of firefly $i$ with firefly $j$ will be potentially useful, the mean of the collaboration scores, denoted by $\mu_{ij}$, must be higher than a certain value $\mu_o$. This statement can be written for hypothesis testing. The objective of this hypothesis testing is, therefore, to check whether $\mu_{ij}$ is greater than or equal to $\mu_o$ with a certain significance level $\alpha$. The problem is left-tailed hypothesis testing, in which the null hypothesis $H_0$ is $\mu_{ij} \geq \mu_o$ and the alternative hypothesis $H_a$ is $\mu_{ij} < \mu_o$. The null hypothesis will be rejected if the value of the test statistic falls in $(-\infty, -t_\alpha^{(n-1)})$. If the null hypothesis is rejected, firefly $i$ will be considered as having ineffective collaborations with firefly $j$ and its impending move will be eliminated. If the null hypothesis is not rejected, firefly $i$ will be allowed to move towards firefly $j$. Since a collaboration score can be either 0 or 1, the mean of the collaboration scores is on $[0,1]$. In this study, the value of $\mu_o$ is set to be a random number on $[0,1]$.

The pseudo-code illustrating the mechanism of the SFA for minimization problems is shown below. In the pseudo-code, $f(\mathbf{x})$ denotes the objective function to be minimized, while $\mathbf{x}_i$ denotes firefly $i$. In addition, $P_{ij}$ represents the hypothesis test result for the collaborations of firefly $i$ with firefly $j$. When $P_{ij} = 1$, it means that the collaborations of firefly $i$ with firefly $j$ are considered effective and the impending move of firefly $i$ can be proceeded.

<u>SFA for minimization problems:</u>

    Generate the initial swarm of $npop$ fireflies;
    Determine $f(\mathbf{x}_i), \ i = 1, \dots, npop$;
    Set the initial number of moves of every firefly ordered pair to 1;
    Give a collaboration score of 1 to every firefly ordered pair;
    Set $P_{ij} = 1, i, j = 1, \dots, npop$;
    $\tau = 0$;
    **While** ($\tau < max\ iteration$)
      **For** $i = 1$ to $npop$
        **For** $j = 1$ to $npop$
          **If** $(f(\mathbf{x}_j) < f(\mathbf{x}_i))$ // firefly $j$ is better than firefly $i$
            **If** $(P_{ij} = 1)$
              Move firefly $i$ towards firefly $j$;
              Determine $f(\mathbf{x}_i)$ and collect the collaboration score;
            **End If**
          Obtain $\mu_0$ randomly on $[0,1]$;
          Perform a hypothesis test to obtain $P_{ij}$;
          **End If**
        **End For**
      **End For**
      Rank the fireflies and find the current best;
      $\tau = \tau + 1$;
    **End While**

## 5. Truss topology optimization

In this study, the objective of topology optimization of a truss is to minimize the weight of the truss by adjusting the cross-sectional areas of its elements. The area of an element can be zero, which simply means that the element is not present. A truss optimization problem considered in this study can be generally expressed as

Minimize

$$W = \sum_{i=1}^{EL} \rho_i L_i A_i \tag{8}$$

Subjected to

Truss is kinematically stable

$$|\sigma_i| \leq |\sigma_{a,i}|, \quad i = 1,2,\ldots,el \tag{9}$$

$$|\delta_j| \leq |\delta_{a,j}|, \quad j = 1,2,\ldots,ndof$$

where

$W$    weight of the truss
$EL$    number of all possible elements from the ground structure
$el$    number of elements from the truss
$ndof$    number of displacement degrees of freedom from the truss
$\rho_i$    weight density of element $i$ of $EL$ elements
$L_i$    length of element $i$ of $EL$ elements
$A_i$    area of element $i$ of $EL$ elements
$\sigma_i$    stress of element $i$ of $el$ elements of the truss
$\sigma_{a,i}$    allowable stress for element $i$
$\delta_j$    displacement degree of freedom $j$ of $ndof$ displacement degrees of freedom of the truss
$\delta_{a,j}$    allowable displacement degree of freedom for displacement degree of freedom $j$.

When a truss satisfies all the constraints, the truss is a feasible truss. Otherwise, the truss is an infeasible one. The optimization problem is Eq. (8) is written for the proposed algorithm as

Minimize

$$f = \begin{cases} W, & \text{if truss is feasible} \\ 10^{20}, & \text{if truss is infeasible.} \end{cases} \quad (10)$$

In some research works [38, 52, 53], large positive values are only assigned to the objective functions of kinematically unstable trusses. Since FAs have a robust mechanism of full-swarm information exchanges, the entire swarm can easily move to feasible regions within some few iterations of the optimization process. Therefore, in this study, the evaluations of search solutions are simplified by assigning a large positive value to the objective function of any infeasible truss.

In addition to the parameters of the constraints in Eq. (9), the values of the minimum area $A_{min}$ and the maximum area $A_{max}$ for each truss member have to be set. If the value of an area design variable becomes lower than $A_{min}$, the area is simply taken as zero. In addition, if the value of an area design variable becomes higher than $A_{max}$, the area is set to $A_{max}$. This kind of decoding is common in truss topology optimization. However, the algorithm in this study does not directly use the topologies obtained from the above simple decoding process. Rather, the obtained topologies are refined by an element-removal algorithm [38] that removes unwanted elements from these topologies to obtain better topologies. Unwanted elements are those elements that are obviously unstable or obviously have no force. The details of the element-removal algorithm can be found in [38].

## 6. Numerical results

The SFA is validated by solving four truss topology optimization problems. These problems are also solved by an FA, whose update equation is presented by Eq. (4). The results obtained by the two algorithms are compared to see how the SFA can reduce the computational efforts shown by running times and numbers of objective function evaluations, and how the SFA can maintain the quality of the obtained results in comparison with those of the FA. The algorithm coefficients are set as follows: $\beta_0 = 2$, $\gamma = 1$, $\omega = 0.978$ and $\theta_{i,k} = $ a random value on [0,1]. In addition, both algorithms are run for 1000 times for each problem to see the quality and uniformity of the obtained results. The significance level of $\alpha = 0.05$ is selected. Each run consists of 1000 iterations. Because of that, the number of moves for a particular ordered pair of fireflies can vary from 1 to 1000. For simplicity, a fixed critical value of $-t_\alpha^{(500-1)} = -1.65$ is used, which is equivalent to having a fixed sample size of 500. In each problem, the FA is run with a swarm of 20 fireflies. However, swarms of 20, 30 and 40 fireflies are used for the SFA in order to investigate the behaviour of the SFA. The following abbreviations, i.e. AMPDE(10×10), SCGA(50), PSO(2×10), FA(20),

SFA(20), SFA(30) and SFA(40), are used, where the numbers in the parentheses indicate the swarm sizes. The programs for both algorithms are coded in C++ and run in a computer with the Intel Core i9-7900X processor and 128 GB RAM.

### 6.1. Problem 1: 12-node, 39-element truss

The first problem is a 2D problem in Fig 6, which has been studied by many researchers [38, 53-55]. The employed ground structure has 12 nodes, and only 39 elements shown in the figure are considered. As the symmetry along the middle vertical line of the ground structure is not assumed in this study, the number of area design variables is 39. Forces, supports, and problem dimensions are also shown in Fig 6. Young's modulus $E = 10^4$ ksi and the weight density $\rho = 0.1$ lb/in$^3$ are employed. In addition, the allowable stress $\sigma_a = \pm\ 20$ ksi and the allowable displacement $\delta_a = \pm\ 2$ in are used. The area $A$ of an element is set to vary from $0.05$ to $2.25$ in$^2$, while the area design variable of an element is set to vary from $-2.25$ to $2.25$ in$^2$. If the area design variable of an element is lower than $0.05$ in$^2$, the area is considered as zero. On the contrary, if the area design variable of an element is greater than $2.25$ in$^2$, the area is set to $2.25$ in$^2$.

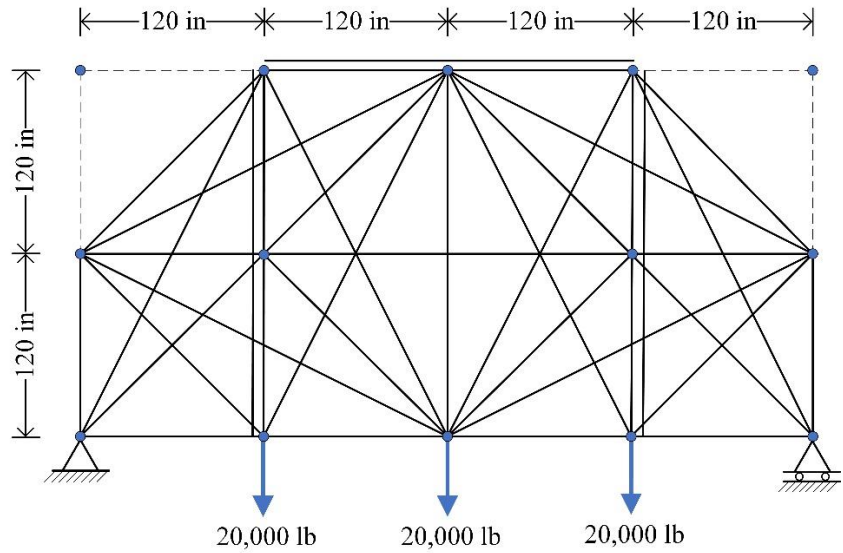

Fig 6. Problem 1: Ground structure

The best topologies obtained by all the cases of the two algorithms are the same and shown in Fig 7. The element areas and minimum weights from the two algorithms are also the same and shown in Table 1. Note that the same best topology is obtained by several researchers [38, 53-55]. In addition, the same minimum weight of 193.200 lb is obtained by many studies [38, 53]. A similar minimum weight of 193.012 lb is obtained in [55], whereas [54] reported a value of 193.547 lb. Note that these research works [38, 53-55] assume the symmetry along the middle vertical line of the ground structure which reduces the number of area design variables from 39 to 21.

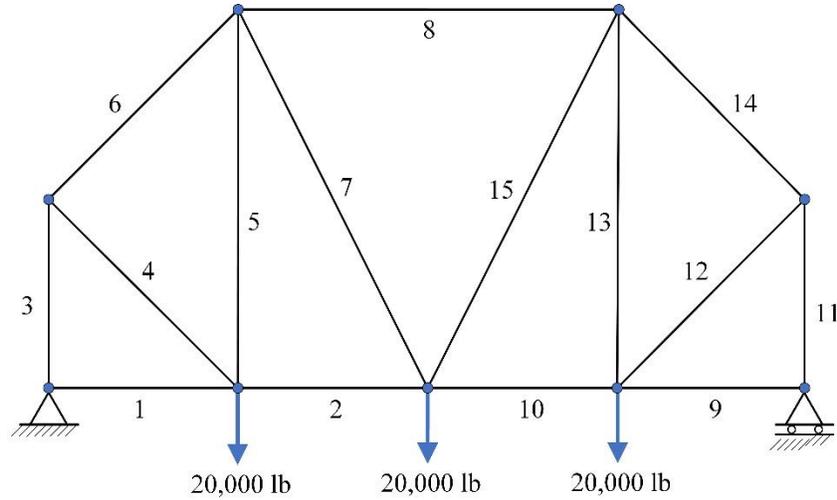

Fig 7. Problem 1: Best topology

Table 1. Problem 1: Element areas of the best solution

| Element | Shakya et al. [38][a] | Wu and Tseng [53][a] | This study |
|---|---|---|---|
| | | Area (in$^2$) | |
| 1,9 | 0.050 | 0.050 | 0.050 |
| 2,10 | 0.750 | 0.750 | 0.750 |
| 3,11 | 1.500 | 1.500 | 1.500 |
| 4,12 | 1.061 | 1.060 | 1.061 |
| 5,13 | 0.250 | 0.250 | 0.250 |
| 6,14 | 1.061 | 1.061 | 1.061 |
| 7,15 | 0.559 | 0.559 | 0.559 |
| 8 | 1.000 | 1.000 | 1.000 |
| Weight (lb) | 193.200 | 193.199 | 193.200 |
| Max stress (ksi) | 20.000 | 20.000 | 20.000 |
| Max displacement (in) | 1.440 | 1.440 | 1.440 |

[a] Symmetry is assumed

In Table 2, the statistics of the obtained results are shown. The table clearly shows that the three cases of the SFA use much smaller computational efforts than the FA, as evidenced by the numbers of objective function evaluations and running times. Although the computational effort for solving the problem by SFA(20) is smaller than that of FA(20), its overall performance is worse than the overall performance of FA(20). However, the overall performance of SFA(30) is comparable to the overall performance of FA(20) while SFA(30) consumes nearly 3.0 times a smaller computational effort. Furthermore, SFA(40) is better than FA(20) in all the statistical results while also requires a smaller computational effort. The best convergences of all the cases are shown in Fig 8. Although the SFA reduces much computation, good convergences, which are comparable to the convergence of the FA, are obtained.

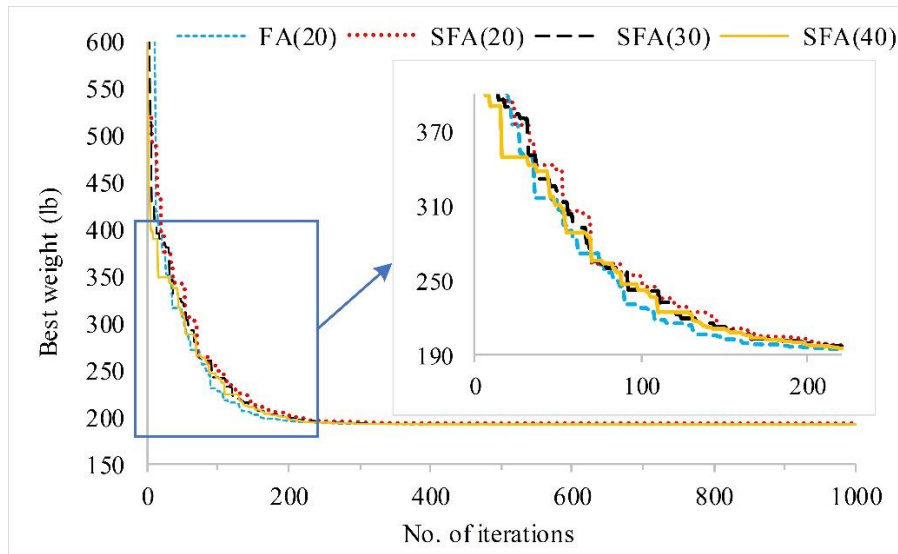

Fig 8. Problem 1: Best convergences of the objective function values

Table 2. Problem 1: Statistics of 1000 run solutions

|  | Shakya et al. [38] | Wu and Tseng [53] | This study | | | |
|---|---|---|---|---|---|---|
| Algorithm | PSO(2×10) | AMPDE(10×10) | FA(20) | SFA(20) | SFA(30) | SFA(40) |
| Minimum weight (lb) | 193.200 | 193.199 | 193.200 | 193.200 | 193.200 | 193.200 |
| Average weight (lb) | – | – | 216.0 | 218.9 | 214.7 | 213.2 |
| Maximum weight (lb) | – | – | 304.6 | 318.8 | 352.3 | 275.6 |
| Standard deviation (lb) | – | – | 15.6 | 18.9 | 15.5 | 14.6 |
| Percentage of the run solutions that are not heavier than the | – | – | 21.5 | 19.2 | 24.2 | 28.4 |

| minimum weight of the entire 1000 runs by greater than 2% (%) | | | | | | |
|---|---|---|---|---|---|---|
| Average number of objective function evaluations | 20000 | – | 189822 | 35242 | 70639 | 116917 |
| Average number of effective collaborations | – | – | 3983 | 3707 | 5338 | 7035 |
| Average running time (sec) | – | – | 32.9 | 7.8 | 14.3 | 22.8 |

*6.2. Problem 2: 10-node, 45-element truss*

The second problem is a 2D problem in Fig 9. Similar to the first problem, this problem has also been studied by many researchers [52, 56-58]. The ground structure, constructed from a 2×5 grid, has 10 nodes. In this study, the symmetry along the middle vertical line of the ground structure is not used. The number of area design variables is equal to 45 since all 45 elements linking all 10 nodes of the ground structure are considered. Forces, supports, and problem dimensions are also shown in Fig 9. Young's modulus $E = 10^4$ ksi and the weight density $\rho = 0.1$ lb/in$^3$ are employed. In addition, the allowable stress $\sigma_a = \pm\,25$ ksi and the allowable displacement $\delta_a = \pm\,2$ in are used. The area $A$ of an element varies from 0.09 to 1.00 in$^2$. The area design variable of an element is set to vary from $-1.00$ to 1.00 in$^2$. If the area design variable of an element is lower than 0.09 in$^2$, the area is considered as zero. If the area design variable of an element is greater than 1.00 in$^2$, the area is set to 1.00 in$^2$.

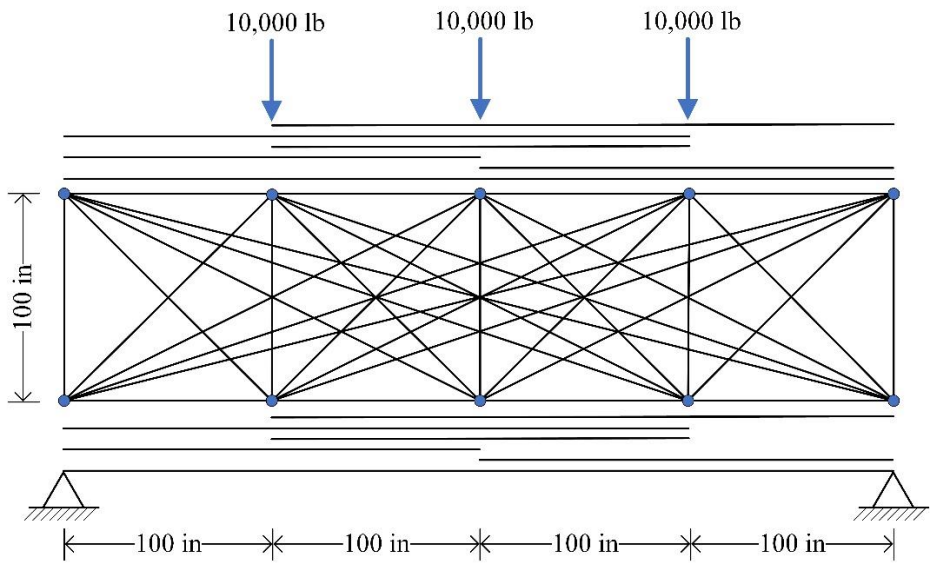

Fig 9. Problem 2: Ground structure

The best topologies obtained by all the cases of the two algorithms are the same and shown in Fig 10. The element areas and minimum weights from the two algorithms are also the same and shown in Table 3. Note that the same best topology is also obtained by several researchers [56-58]. In addition, the same minimum weight of 44.000 lb is obtained in [56, 57] and a similar minimum weight of 44.033 lb is obtained in [52].

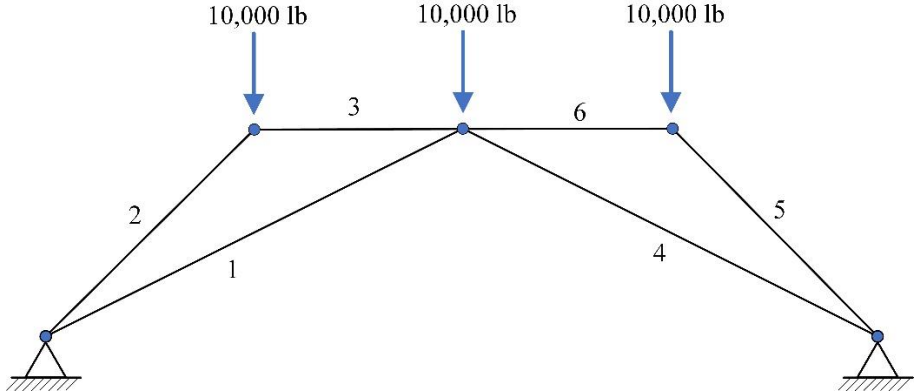

Fig 10. Problem 2: Best topology

Table 3. Problem 2: Element areas of the best solution

|  | Faramarzi and Afshar [56] | Li [57] | This study |
|---|---|---|---|
| Element | Area (in$^2$) | | |
| 1 | 0.447 | 0.447 | 0.447 |
| 2 | 0.566 | 0.566 | 0.566 |
| 3 | 0.400 | 0.400 | 0.400 |
| 4 | 0.447 | 0.447 | 0.447 |
| 5 | 0.566 | 0.566 | 0.566 |
| 6 | 0.400 | 0.400 | 0.400 |
| Weight (lb) | 44.000 | 44.000 | 44.000 |
| Max stress (ksi) | 25.000 | 25.000 | 25.000 |
| Max displacement (in) | 1.250 | 1.250 | 1.250 |

In Table 4, the statistics of the obtained results are shown. Again, the three cases of the SFA use much smaller computational efforts than the FA. The computational effort of solving the problem with SFA(20) is about 6 times smaller than that of FA(20). However, its overall performance is only slightly worse than that of FA(20). In addition, the SFA(30) performs slightly better than FA(20) while taking about 3 times a smaller computational effort. Furthermore, SFA(40) also consumes a smaller computational effort than FA(20), while surpasses FA(20) in all the statistical results. The best convergences of all the cases are shown in Fig 11. With smaller computational efforts, the SFA still gives good convergences which are comparable to the convergence given by the FA.

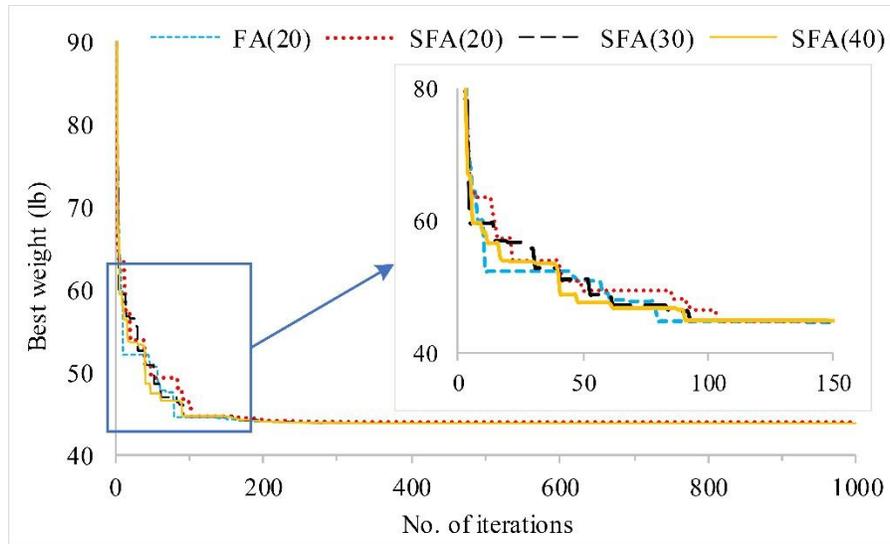

Fig 11. Problem 2: Best convergences of the objective function values

Table 4. Problem 2: Statistics of 1000 run solutions

|  | Faramarzi and Afshar [56] | Li [57] | This study | | | |
|---|---|---|---|---|---|---|
| Algorithm | CA-LP | SCGA(50) | FA(20) | SFA(20) | SFA(30) | SFA(40) |
| Minimum weight (lb) | 44.000 | 44.000 | 44.000 | 44.000 | 44.000 | 44.000 |
| Average weight (lb) | – | – | 44.4 | 44.6 | 44.3 | 44.1 |
| Maximum weight (lb) | – | – | 64.9 | 91.6 | 60.9 | 52.0 |
| Standard deviation (lb) | – | – | 1.1 | 2.4 | 0.8 | 0.4 |
| Percentage of the run solutions that are not heavier than the minimum weight of the entire 1000 runs by greater than 2% (%) | – | – | 68.8 | 63.1 | 72.1 | 86.5 |
| Average number of objective function evaluations | – | – | 190095 | 31284 | 64157 | 107567 |

| Average number of effective collaborations | – | – | 2492 | 2230 | 3396 | 4581 |
| Average running time (sec) | – | – | 42.5 | 7.0 | 14.5 | 24.4 |

*6.3. Problem 3: 35-node, 595-element truss*

The third problem is a 2D problem in Fig 12 [56]. In this study, the number of nodes in the ground structure is increased from 12, used in [56], to 35 in Fig 12. Note that the grid spacing is uniform. All 595 elements of full connections between 35 nodes of the ground structure are considered. The symmetry along the middle horizontal line of the ground structure is assumed. As a result, the number of area design variables is reduced from 595 to 315. Force, supports, and problem dimensions are also shown in Fig 12. Young's modulus $E = 10^4$ ksi and the weight density $\rho = 0.1$ lb/in$^3$ are employed. In addition, the allowable stress $\sigma_a = \pm 5$ ksi and the allowable displacement $\delta_a = \pm 0.6$ in are used. The area $A$ of an element is set to vary from 0.09 to 35.00 in$^2$, while the area design variable of an element is set to vary from $-35.00$ to 35.00 in$^2$. If the area design variable of an element is lower than 0.09 in$^2$, the area is considered as zero. If the area design variable of an element is greater than 35.00 in$^2$, the area is set to 35.00 in$^2$.

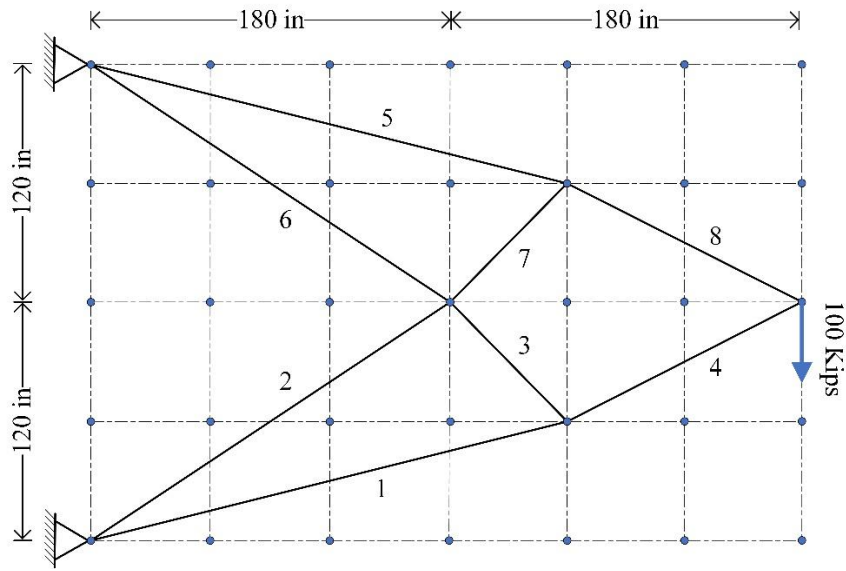

Fig 12. Problem 3: Ground structure and best topology

The best topologies obtained by all the cases of the two algorithms in this study are the same and shown in Fig 12. The element areas and minimum weights from the two algorithms are also the same and shown in Table 5. The minimum weight of 2232.000 lb is better than the result obtained in [56], which is 2400.001 lb.

Table 5. Problem 3: Element areas of the best solution

| Element | Area (in$^2$)[a] |
|---|---|
| 1, 5 | 24.739 |
| 2, 6 | 7.211 |
| 3, 7 | 5.657 |
| 4, 8 | 22.361 |
| Weight (lb) | 2232.000 |
| Max stress (ksi) | 5.000 |
| Max displacement (in) | 0.558 |

[a] Symmetry is assumed

In Table 6, the statistics of the obtained results are shown. It is found that the three cases of the SFA use much smaller computational efforts than the FA. Although the computational effort to solve the problem with FA(20) is much larger than those of SFA(20) and SFA(30), its overall performance is better than the overall performances of SFA(20) and SFA(30). In addition, SFA(40) consumes about 1.5 times a smaller computational effort than FA(20) while it still performs slightly better than FA(20) in all the statistical results. The best convergences of all the cases are shown in Fig 13. Although the SFA consumes much smaller computational efforts than the FA, the SFA can yield good convergences that are comparable to that given by the FA.

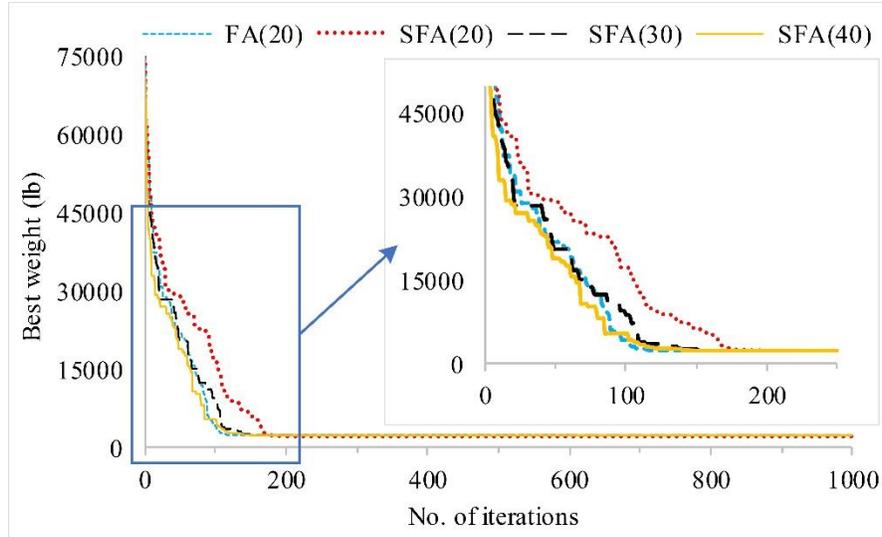

Fig 13. Problem 3: Best convergences of the objective function values

Table 6. Problem 3: Statistics of 1000 run solutions

|  | Faramarzi and Afshar [56] | This study | | | |
| --- | --- | --- | --- | --- | --- |
| Algorithm | Cellular automata | FA(20) | SFA(20) | SFA(30) | SFA(40) |
| Minimum weight (lb) | 2400.0 | 2232.0 | 2232.0 | 2232.0 | 2232.0 |
| Average weight (lb) | − | 2766.8 | 3281.4 | 2920.1 | 2757.4 |
| Maximum weight (lb) | − | 4640.2 | 5785.7 | 5039.7 | 4568.1 |
| Standard deviation (lb) | − | 420.3 | 642.5 | 489.1 | 412.9 |
| Percentage of the run solutions that are not heavier than the minimum weight of the entire 1000 runs by greater than 2% (%) | − | 2.3 | 0.6 | 1.3 | 2.6 |
| Average number of objective function evaluations | − | 189338 | 41492 | 74496 | 117923 |
| Average number of effective collaborations | − | 6341 | 5112 | 5057 | 5608 |
| Average running time (sec) | − | 467.2 | 116.6 | 200.5 | 317.8 |

*6.4. Problem 4: 18-node, 153-element truss*

The last problem is a 3D problem in Fig 14, which is studied in [38]. The ground structure has 18 nodes and is constructed by a 3×3×2 grid. All 153 elements of full connections between 18 nodes are considered. The number of area design variables is 153 since the symmetry is not assumed. Force, supports, and problem dimensions are also shown in Fig 14. Young's modulus $E = 200$ GPa and the weight density $\rho = 77$ kN/m$^3$ are employed. In addition, the allowable stress $\sigma_a = \pm 150$ MPa and the allowable displacement $\delta_a = \pm 20$ mm are used. The area $A$ of an element is set to vary from 0 to 2000 mm$^2$, while the area design variable of an element is set to vary from $-2000$ to 2000 mm$^2$. If the area design variable of an element is less than or equal to 0 mm$^2$, the area is considered as zero. On the contrary, if the area design variable of an element is greater than 2000 mm$^2$, the area is set to 2000 mm$^2$.

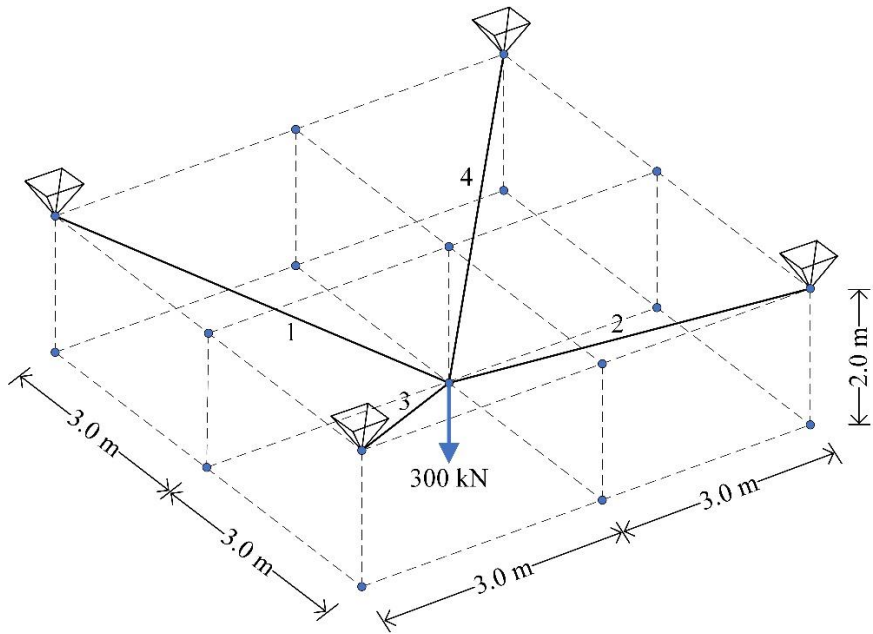

Fig 14. Problem 4: Ground structure and best topology

The best topologies obtained by all the cases of the two algorithms are the same and shown in Fig 14. The minimum weights from the two algorithms are also the same and equal to 1694.0 N. Among the best solutions, the areas of all four members are not always identical. However, the areas of elements 1 and 2 are always identical, and the areas of elements 3 and 4 are always identical. Note that the same best topology and minimum weight of 1694.0 N are also obtained in [38].

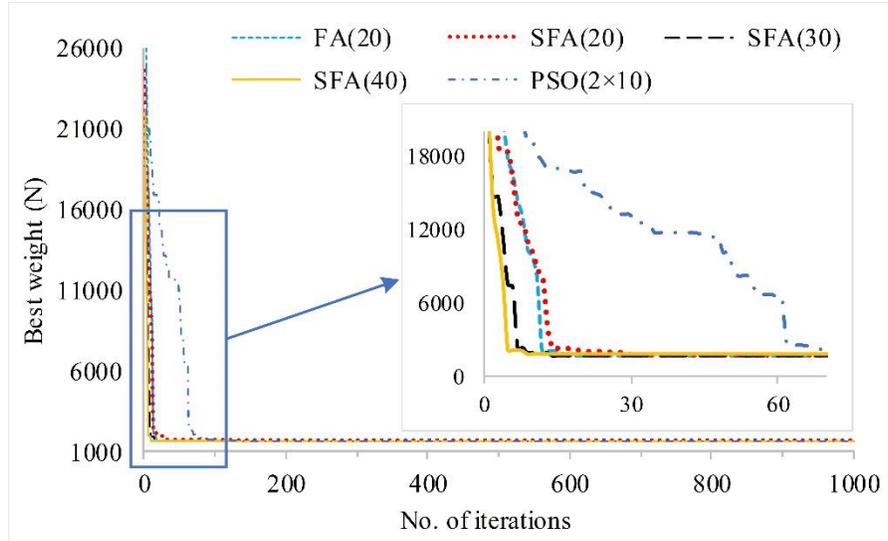

Fig 15. Problem 4: Best convergences of the objective function values

Table 7. Problem 4: Statistics of 1000 run solutions

|  | Shakya et al. [38] | This study | | | |
|---|---|---|---|---|---|
| Algorithm | PSO(2×10) | FA(20) | SFA(20) | SFA(30) | SFA(40) |
| Minimum weight (N) | 1694.0 | 1694.0 | 1694.0 | 1694.0 | 1694.0 |
| Average weight (N) | 2194.8 | 1751.1 | 1774.1 | 1707.3 | 1695.8 |
| Maximum weight (N) | 8976.5 | 3738.9 | 5756.8 | 2910.5 | 2442.3 |
| Standard deviation (N) | 986.2 | 239.4 | 313.3 | 114.2 | 33.6 |
| Percentage of the run solutions that are not heavier than the minimum weight of the entire 1000 runs by greater than 2% (%) | 69.6 | 93.8 | 92.5 | 98.6 | 99.7 |
| Average number of objective function evaluations | 20000 | 189910 | 31018 | 62663 | 104774 |
| Average number of effective collaborations | - | 1950 | 1930 | 2763 | 3643 |
| Average running time (sec) | - | 179.4 | 30.1 | 61.1 | 102.1 |

In Table 7, the statistics of the obtained results are shown. It can be observed that the statistical results by the FA and the SFA are much better than the results obtained in [38], in which a multi-population PSO [59-61] is employed. Although FA(20) has a better performance than PSO(2×10), it also requires much larger computational effort. With a reasonable average number of objective function evaluations, SFA(20) performs much better than PSO(2×10). Table 7 clearly shows that the SFA can dramatically reduce the computational efforts of the conventional FA. The computational effort of solving the problem by SFA(20) is nearly 6 times smaller than that of FA(20). However, its overall performance is worse than the performance of FA(20). Nevertheless, the difference is insignificant. In addition, SFA(30) and SFA(40) perform much better than FA(20) in all the statistical results while also consume smaller computational efforts. The best convergences of all the cases are shown in Fig 15. Although the SFA consumes smaller computational efforts than the FA, good convergences are still obtained, which are comparable to the convergence given by the FA. In addition, the convergences obtained by the FA and the SFA are much faster than the one given by a multi-population PSO in [38].

## 7. Conclusions

In this study, a statistical firefly algorithm (SFA) is proposed. In the proposed SFA, mean hypothesis testing is applied to the mechanism of an ordinary FA during the optimization process in order to limit the movements of fireflies to only those that are potential useful. To be able to predict the potential usefulness of the collaborations between firefly couples, the quality of their collaborations is continuously collected by considering the results of their movements. After that, the statistics of these results are computed. To accept that any collaboration between an ordered pair of fireflies will be useful, the mean score of their collaborations must be higher than a certain value. This statement is written for hypothesis testing, in which the statistics of the results of the past collaborations are utilized. Only when each ordered pair of fireflies passes each hypothesis test, can their current collaboration be proceeded.

The proposed SFA is used to solve four truss topology optimization problems to demonstrate its validity and efficiency. After the statistical strategy is added, a large number of fireflies' movements are blocked since potentially ineffective collaborations are identified. Subsequently, the SFA takes significantly less computational efforts than the FA to yield similarly accurate results. In addition, this study also shows that the FA and the SFA have better performances than PSO. While the FA performs better than PSO with larger computational efforts, the SFA can have better performances than PSO with reasonable computational efforts. The SFA can also yield fast convergences of objective function values. Therefore, the SFA can be considered as a robust optimization tool for tackling large and complex problems that require large computational efforts.

**Acknowledgments**

Scholarships from the ASEAN University Network/Southeast Asia Engineering Education Development Network (AUN/SEED-Net) and Sirindhorn International Institute of Technology for the first author are greatly appreciated.